\documentclass[letterpaper]{article} 
\usepackage{aaai2026}  
\usepackage{times}  
\usepackage{helvet}  
\usepackage{courier}  
\usepackage[hyphens]{url}  
\usepackage{graphicx} 
\usepackage{natbib}  
\usepackage{caption} 
\usepackage{algorithm}
\usepackage{algorithmic}

\usepackage{amsmath}   
\usepackage{newtxmath}
\usepackage{longtable}
\usepackage{booktabs}   
\usepackage{multirow}
\usepackage{rotating}
\usepackage{array}          
\usepackage{ragged2e} 
\usepackage{pifont}
\newcommand{\cmark}{\ding{51}} 

\usepackage[caption=false]{subfig}  
\usepackage{newfloat}
\usepackage{listings}
\DeclareCaptionStyle{ruled}{labelfont=normalfont,labelsep=colon,strut=off} 
\floatstyle{ruled}
\newfloat{listing}{tb}{lst}{}
\floatname{listing}{Listing}
\title{Fine-Grained DINO Tuning with Dual Supervision \\for Face Forgery Detection}
\author{
    Tianxiang Zhang, Peipeng Yu, Zhihua Xia\thanks{Corresponding author: Zhihua Xia .}, Longchen Dai, Xiaoyu Zhou, Hui Gao\\
}
\affiliations{
    College of Cyber Security, Jinan University\\

    \{11027ztx, gaohui\}@stu.jnu.edu.cn, \{ypp865, xia\_zhihua\}@163.com, longchendai@stu2023.jnu.edu.cn, xiaoyuzhou68@stu2024.jnu.edu.cn
}

\usepackage{bibentry}

\begin{document}

\maketitle

\begin{abstract}
The proliferation of sophisticated deepfakes poses significant threats to information integrity. While DINOv2 shows promise for detection, existing fine-tuning approaches treat it as generic binary classification, overlooking distinct artifacts inherent to different deepfake methods. To address this, we propose a DeepFake Fine-Grained Adapter (DFF-Adapter) for DINOv2. Our method incorporates lightweight multi-head LoRA modules into every transformer block, enabling efficient backbone adaptation. DFF-Adapter simultaneously addresses authenticity detection and fine-grained manipulation type classification, where classifying forgery methods enhances artifact sensitivity. We introduce a shared branch propagating fine-grained manipulation cues to the authenticity head. This enables multi-task cooperative optimization, explicitly enhancing authenticity discrimination with manipulation-specific knowledge. Utilizing only 3.5M trainable parameters, our parameter-efficient approach achieves detection accuracy comparable to or even surpassing that of current complex state-of-the-art methods.
\end{abstract}


\section{Introduction}

In recent years, the explosive growth of deep-fake technology has ushered in a revolutionary breakthrough in multimedia content generation, dramatically extending the practical boundaries of visual synthesis~\cite{survey1}.The misuse of deepfake technology raises profound security concerns, eroding trust through identity theft, privacy violations, and misinformation.With the rapid advancement of generative models, synthetic content increasingly bypasses traditional forensic methods, making the development of broadly generalisable detection systems an urgent priority in AI security~\cite{survey2,survey3}.

To address the generalization challenges in deepfake detection, existing approaches can be broadly categorized into four types: detection based on physiological/physical artifacts~\cite{lips}, noise residual analysis~\cite{frepgan}, feature consistency analysis~\cite{finfer}, and pre-trained large models based detection~\cite{uia}. Recent research have increasingly adopted pretrained large model to extract more expressive and semantically rich feature representations. Fine-tuning methods~\cite{LoRA,multihead} based on the ViT architecture have achieved state-of-the-art performance on multiple public deepfake datasets, validating their broad applicability and remarkable potential in forgery detection tasks, while offering a promising pathway to enhance model generalization~\cite{RepDFD}. Meanwhile, the emergence of recent benchmarks featuring diverse generative models and demographic variations~\cite{AI-Face} and ILLUSION~\cite{ILLUSION} further emphasizes the need for detectors with stronger generalization and fairness. However, existing detection methods based on pre-trained large models either insert an adapter in the final Transformer block or apply LoRA to fine-tune parameters. These approaches lack task-specific design for deepfake detection and struggle to surpass the upper bound of generalization performance.

\begin{figure}[t]
  \centering
  \includegraphics[width=\columnwidth]{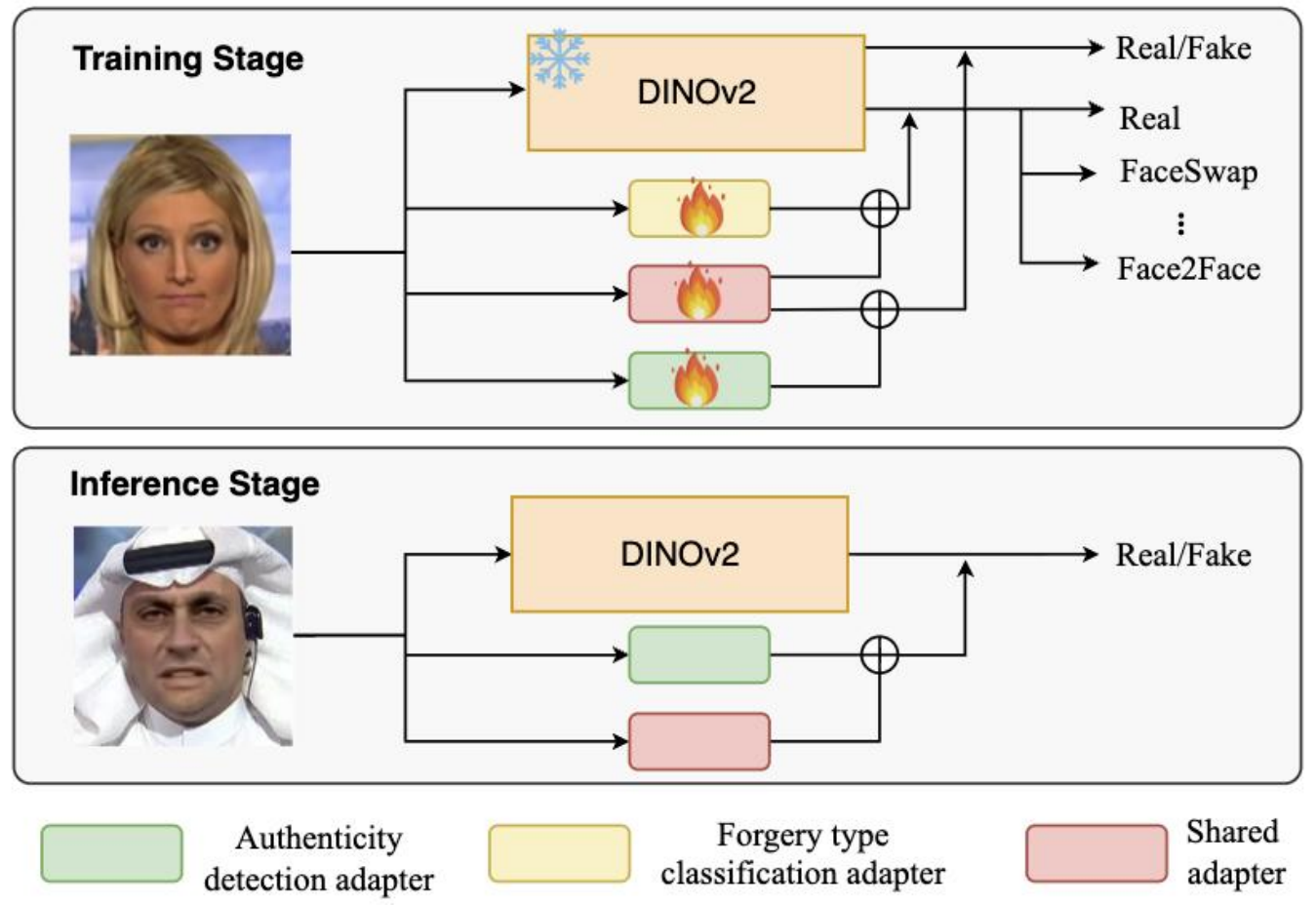}
  \caption{\textbf{Training and Inference Stages.} During training, the frozen DINOv2 backbone with the DFF-Adapter is augmented by three adapter heads: authenticity, forgery-type, and shared. The authenticity and forgery-type branches are jointly optimized, while the shared branch captures fine-grained forgery cues and transfers them to the authenticity stream. During inference, only the fused authenticity and shared branches are used for face forgery detection.}
  \label{fig:overview}
\end{figure}

To address this issue, we propose the DeepFake Fine-Grained Adapter (DFF-Adapter), a fine-grained tuning method for deepfake detection based on DINOv2. Specifically, DFF-Adapter is integrated into every Transformer block of DINOv2 to enable the joint optimization of two tasks: authenticity detection and forgery type classification.
The architecture comprises three branches: an authenticity detection head, a forgery type classification head, and a shared head adapter. The shared head participates in feature modeling for both tasks, aiming to effectively transfer the fine-grained forgery cues extracted by the forgery type branch to the authenticity branch. This allows the main task to go beyond relying solely on coarse global signals and instead benefit from the detailed artifact patterns learned by the auxiliary task, leading to more generalizable forgery detection (see in Figure.~\ref{fig:overview}).
Additionally, we divide the input features of each adapter head into multiple subspaces and introduce a multi-head composition mechanism that enables different subspaces to focus on distinct aspects of forgery-related features, thereby achieving multi-view feature fusion. 

Our main contributions are summarized as follows:
\begin{itemize}
    \item We devise a DeepFake Fine-Grained Adapter architecture that intertwines a shared branch with task-specific branches, enabling the authenticity detector to inherit fine-grained forgery cues from the forgery-type classification task. 
    \item We propose a Forgery-Aware Multi-Head Router that partitions intermediate Transformer features into multiple subspaces and, for each subspace, dynamically routes to a learned top-3 set of LoRA experts. This per-subspace expert optimization fully mines localized forgery artefacts and enables fine-grained, multi-view feature fusion.
    \item Our method achieves superior generalization in cross-dataset evaluations, outperforming state-of-the-art approaches on multiple challenging benchmarks. Furthermore, in cross-manipulation evaluations conducted on the recently proposed DF40 dataset, it also achieves the best overall performance among all competing methods.
\end{itemize}

\begin{figure*}[t]
  \centering
  \includegraphics[width=\linewidth]{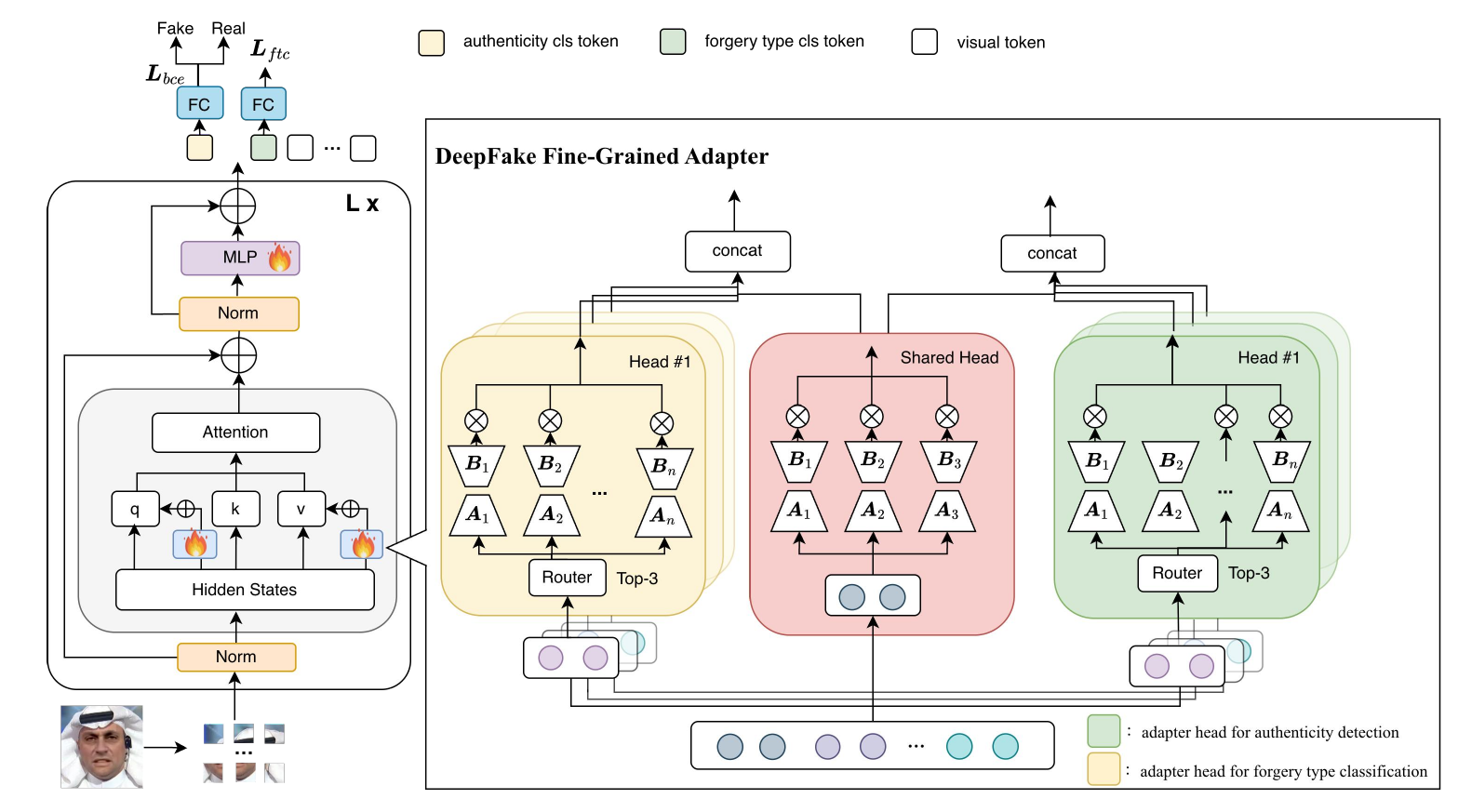}
  \caption{The framework of our method augments a frozen DINOv2 backbone with DFF-Adapters placed in each Transformer block. Each adapter contains three low-rank heads—authenticity, forgery-type, and shared—whose multi-head routers select the top-3 LoRA experts per feature subspace. The shared head transfers fine-grained cues to the authenticity stream. During training, the authenticity and forgery-type CLS tokens are supervised by a binary cross-entropy loss \(L_{\text{bce}}\) and a multi-class loss \(L_{\text{ftc}}\), respectively.}
  \label{fig:arch}
\end{figure*}

\section{Related Work}

\subsection{Classical Forgery Detection Methods}
To improve the generalization ability of face forgery detection models against unseen manipulation methods, researchers have proposed numerous detection methods.Early research concentrated on identifying physiological or physical artifacts introduced during the synthesis process, targeting visually observable inconsistencies directions~\cite{lips,fwa}. However, with the rapid advancement of generative models, these artifacts have become increasingly subtle and less discernible. In response, researchers have proposed noise residual analysis methods that exploit subtle anomalies in the frequency domain~\cite{frepgan,dynamic,frequency}; Methods based on feature consistency focus on detecting latent inconsistencies among internal representations within forged images, capturing contradictions between manipulated and authentic regions~\cite{lec,wsc,delocate}.In contrast to these traditional techniques, recent methods based on pre-trained large models leverage powerful visual representations learned from diverse, large-scale datasets. These models exhibit superior global perception, semantic abstraction, and robust generalization to unseen forgeries. pre-trained large models offer a unified and extensible framework for face forgery detection across diverse manipulation techniques and data domains.

\subsection{Pre-trained Large Models for Detection}
Methods based on pre-trained large models leverage knowledge distilled from vast image corpora and fine-tune deep networks for the forgery detection task. Prior studies have explored various architectural improvements to the Vision Transformer~\cite{exploring}, and some research focuses on uncovering the visual priors embedded in clip~\cite{LVLM-DFD,VB-StA}. By employing lightweight fine-tuning strategies, these methods have effectively enhanced the model’s generalization ability in cross-dataset scenarios.

DINO is a visual foundation model based on Vision Transformer~\cite{dino}. It employs self-supervised knowledge distillation to learn structure-aware and highly generalizable visual features from large-scale unlabeled image datasets, thereby encoding rich prior knowledge. In deepfake detection tasks, simply appending a linear classifier to DINOv2 can significantly outperform supervised models such as DeiT-III and CLIP during testing~\cite{expss}. Compared to CLIP, which operates with dual image-text branches and focuses on semantic alignment, DINOv2 adopts a purely visual self-supervised learning paradigm, preserving fine-grained local textures and geometric structures more effectively. It has shown superior performance in dense prediction tasks such as semantic segmentation and patch matching compared to weakly-supervised models like OpenCLIP~\cite{DINOv2}, making it more sensitive to subtle forgery traces in images.

Current detection methods based on DINOv2 commonly insert adapters only into the final Transformer block or fine-tune merely the last few layers~\cite{truthlens,pudd}. However, such a limited fine-tuning scope constrains the backward propagation of task-specific signals to earlier layers, thereby hindering the modeling of low-level forgery cues and lacking specificity for the deepfake detection task.To address these limitations, we propose the DeepFake Fine-Grained Adapter (DFF-Adapter), which injects both task-specific and shared low-rank adapters throughout the entire DINOv2 backbone. This design is tailored specifically for face forgery detection, aiming to capture the distinctive artifact patterns characteristic of various manipulation methods. By modeling fine-grained forgery cues through task-aware adaptation, DFF-Adapter enhances the model’s sensitivity to manipulation-specific traces and effectively improves the generalization capability of deepfake detection approaches.

\section{DeepFake Fine-Grained Adapter}

\subsection{Overview}
Vision Transformers such as DINOv2 have recently been adopted for forgery detection and already outperform conventional CNNs. Most existing methods treat deepfake detection as a binary classification task and fine-tune only the final layers of the backbone accordingly. However, such methods overlook the fact that different forgery methods often generate distinct artefactual patterns. These method-specific cues are informative yet underutilized, limiting the model’s ability to generalize across manipulation types.

To address this limitation, we propose the DeepFake Fine-Grained Adapter (DFF-Adapter), a lightweight, low-rank tuning scheme tailored for deepfake detection. DFF-Adapter jointly optimizes authenticity discrimination and forgery-type classification by injecting task-specific and shared adapters into every Transformer block of a frozen DINOv2 backbone. To further enhance feature diversity and generalization, we design two core modules: a Forgery-Aware Multi-Head Router, which captures diverse forgery artefacts by adaptively routing feature subspaces to specialized experts, and a Shared-Enhanced Task Fusion module, which integrates multi-level task-specific and shared representations to transfer fine-grained forgery cues from the auxiliary branch, thereby enhancing the authenticity detector’s sensitivity to manipulation artefacts.

\subsection{Forgery‑Aware Multi‑Head Router}
To capture the diverse forgery artefacts that manifest in different channel sub-spaces, 
we propose a lightweight DeepForgery Multi‑Head Router(DF-MHR) into our DeepFake Fine-Grained Adapter. 
DF-MHR partitions the input feature map along the channel axis into $h$ disjoint heads, 
enabling independent analysis of each sub-space. 
All head adapters share a pool of $N$ low-rank LoRA adapters; 
for every head, a router scores the adapters and activates the top-3 that best fit the statistics of that sub-space. 
This per-sub-space expert allocation equips each head with a specialised adapter set tuned to the artefacts present in its own feature slice. 
Finally, the head adapter outputs are concatenated into a unified representation that strengthens forgery modelling 
while introducing only a modest number of parameters.

Given the hidden states of a Transformer block
\(\mathbf{X}\in\mathbb{R}^{L\times d}\),
we split the channel axis into \(h\) head adapters,
\(\mathbf{X}^{(k)}\in\mathbb{R}^{L\times d_h}\) with \(d_h=d/h\).
The router keeps two routing-logit tables: a task-specific tensor $\mathbf{Z}_{\mathrm{task}}\!\in\!\mathbb{R}^{T\times h_t\times N}$ serving the $h_t$ task head adapters, and a global vector $\mathbf{Z}_{\mathrm{shared}}\!\in\!\mathbb{R}^{N}$ used by the shared head adapter.
\paragraph{Task-Specific Head Adapter}  
Each task-specific head adapter draws from a shared bank of $N$ low-rank LoRA experts $\{(\mathbf{A}_j,\mathbf{B}_j)\}_{j=1}^{N}$. For a slice width $d_h$, the two projection matrices in expert $j$ have shapes $d_h\times r_h$ and $r_h\times d_h$, where the per-head rank is $r_h=r/h$.  For each task-specific head adapter \(k=1,\dots,h_t\) under task \(t\), we compute the output feature \({f}_{t}^{k}\) as follows:
\begin{equation}
{f}_{t}^{k} =
  \beta \sum_{j\in S_{t,k}}
  \tilde{g}^{(j)}_{t,k}\,
  \mathbf{B}_j\bigl(\mathbf{A}_j\mathbf{X}^{(k)}\bigr) \label{eq:ftk} \\[2pt]
\end{equation}
\begin{equation}
\tilde{g}^{(j)}_{t,k} =
  \frac{g^{(j)}_{t,k}}
       {\sum_{\ell\in S_{t,k}}g^{(\ell)}_{t,k}}, \quad
  \label{eq:alpha} \\[2pt]
\end{equation}
\begin{equation}
S_{t,k} =
\operatorname{Top}\!3\, (N, {g}_{t,k}) \label{eq:top3} \\[2pt]
\end{equation}
\begin{equation}
{g}_{t,k} =
\sigma\!\ \bigl(\,\mathbf{Z}_{\mathrm{task}}[t,k]\,) \label{eq:softmax}
\end{equation}
\noindent
where \(\sigma(\cdot)\) denotes the softmax function. \(\operatorname{Top}3(N,\cdot)\) denotes the set comprising the indices of the three highest routing probabilities among those computed for head adapter \(k\) of task \(t\) across all \(N\) LoRA experts. \({g}_{t,k}\) is the gating scores with head adapter \(k\) of task \(t\).
\(\mathbf{Z}_{\mathrm{task}}[t,k]\) is the score for the \(N\) LoRA experts associated, and
\(\beta\) is a scaling coefficient controlling the magnitude of aggregated LoRA expert outputs.

\paragraph{Shared Head Adapter}  
Given the global logit vector \(\mathbf{Z}_{\mathrm{shared}}\in\mathbb{R}^{N}\),
the shared head  adapter output is computed in a single expression:
\begin{equation} \label{eq:shared_head}
f_{\mathrm{share}}
  = \beta \sum_{j=1}^{N}
   \sigma\!\bigl(\mathbf{Z}_{\mathrm{shared}}\bigr)_{j}\,
    \mathbf{B}_j\bigl(\mathbf{A}_j\mathbf{X}^{(0)}\bigr),
\end{equation}
where \(\mathbf{X}^{(0)}\) is the feature slice for the shared head adapter and \(\beta\) is a scaling coefficient controlling the magnitude of aggregated LoRA expert outputs.

\subsection{Shared-Enhanced Task Fusion}
To improve generalization under diverse manipulation techniques, this module enhances task-aware representation learning by residually fusing shared and task-specific updates across all Transformer blocks, thereby enabling the authenticity detector to inherit auxiliary task knowledge and capture manipulation-specific cues.

The DFF-Adapter inserts low-rank updates into every query, value, and dense projection of each Transformer block. At block \(l\) it yields one shared update
\({f}^{(l)}_{\mathrm{share}}\) and \(n\) task-specific updates
\(\{{f}^{(l),k}_{i}\}_{k=1}^{n}\), where \(i\in\{0,1\}\) indexes authenticity and forgery-type tasks, respectively.

Across the \(L\) blocks, these updates are added residually to the token stream, so the CLS token \({f}_{\mathrm{bin}}\) can be generally expressed as:
\begin{equation}
{f}_{\mathrm{bin}}
  = {h}_{\mathrm{cls}}
  + \operatorname{concat}(
        {f}_{\mathrm{share}},
        {f}^{1}_{0},\dots,{f}^{n}_{0} 
    \bigr) ,
\end{equation}
where \({h}_{\mathrm{cls}}\) is the CLS token from the frozen
backbone, and \(\operatorname{concat}(\cdot)\) stacks the shared output
\({f}_{\mathrm{share}}\) with the \(n\) authenticity-specific
outputs \(\{{f}^{k}_{0}\}\) along the channel dimension.Providing \({f}_{\mathrm{bin}}\) as input to the binary classifier,
the binary-cross-entropy loss \(\mathcal{L}_{\mathrm{bce}}\) is formulated as follows:
\begin{equation}
\mathcal{L}_{\mathrm{bce}} = \mathrm{BCE}\big(y,\, h_{\mathrm{bce}}({f}_{\mathrm{bin}})\big),
\end{equation}
where $\mathrm{BCE}(\cdot, \cdot)$ denotes the binary cross-entropy loss function that measures the discrepancy between the predicted probability and the ground truth label, $h_\mathrm{bce}(\cdot)$ denotes the binary classification layer that converts its input into the predicted probability of the fake class,y is the ground truth label.

Similarly, the CLS token for the forgery-type branch is
\begin{equation}
{f}_{\mathrm{ftc}}
  = {h}_{\mathrm{cls}}
  + \operatorname{concat}(
        {f}_{\mathrm{share}},
        {f}^{1}_{1},\dots,{f}^{n}_{1}
    \bigr),
\end{equation}
where \(\operatorname{concat}(\cdot)\) stacks the shared output
\(f_{\mathrm{share}}\) with the \(n\) forgery-type–specific
outputs \(\{f^{k}_{1}\}\) along the channel dimension.Providing $f_{\mathrm{ftc}}$ as input to the forgery type
classifier, the forgery type classification loss $\mathcal{L}_{\mathrm{ftc}}$ is formulated as follows:
\begin{equation}
\mathcal{L}_{ftc}
  = -\sum_{c=1}^{C} y_{c}\,
      \log h_{\text{ftc}}(f_\mathrm{ftc}),
\end{equation}
where \( y_c\) represents the integer-encoded ground-truth label indicating the correct forgery type among \( C \) categories,and   
the function \( h_{\mathrm{ftc}}(\cdot) \) denotes the predicted probability for class \( c \).

\subsection{Training Details}
To fully leverage the complementary strengths of authenticity discrimination and forgery-type recognition, we adopt a dual-branch training strategy where each task is optimized in a separate forward pass within the same mini-batch. This task-wise decoupling avoids gradient interference, allowing each branch to focus on learning task-specific patterns. Meanwhile, the shared adapter facilitates cross-task knowledge transfer, enabling the authenticity branch to benefit from the fine-grained forgery cues captured by the auxiliary classification task.

 During each mini-batch, the frozen backbone processes the same input images twice.
In the first forward pass, the task flag is set to zero to indicate the authenticity prediction task, and the network produces the authenticity logits. In the second forward pass, the task flag is switched to one to indicate the forgery-type classification task, and the network outputs the forgery-type logits. The total loss is defined as:
\begin{equation}
\mathcal{L}
  = \lambda_{0}\,\mathcal{L}_{\mathrm{bce}}
    + \lambda_{1}\,\mathcal{L}_{\mathrm{ftc}},
\label{eq:total_loss}
\end{equation}
where \(\lambda_0\) and \(\lambda_1\) are weighting hyperparameters that control 
the relative contributions of the authenticity and forgery-type branches. This total loss is back-propagated to update only the DeepFake Fine-Grained Adapters and task-specific classifiers, while keeping all backbone parameters frozen.

\begin{table*}[t]
\centering
\small
\setlength{\tabcolsep}{4pt}
\renewcommand{\arraystretch}{1.25}

\begin{tabular}{
    c|>{\centering\arraybackslash}p{1.6cm}|c|
    >{\centering\arraybackslash}p{1.1cm}
    >{\centering\arraybackslash}p{1.1cm}
    >{\centering\arraybackslash}p{1.1cm}
    >{\centering\arraybackslash}p{1.1cm}
    >{\centering\arraybackslash}p{1.1cm}
}
\toprule
\multirow{2}{*}{\centering \textbf{Method}} & 
\multirow{2}{*}{\centering \textbf{Venue}} & 
\multicolumn{1}{c|}{\textbf{Intra-dataset}} & 
\multicolumn{5}{c}{\textbf{Cross-dataset}} \\
\cline{3-8}
& & \textbf{FF++} & \textbf{CDF-v2} & \textbf{DFDC} & \textbf{CDF-v1} & \textbf{DFDCP} & \textbf{Avg.} \\
\midrule
Xception~\cite{ff++} & ICCV’19  & 97.23 & 81.65 & -- & 80.98 & 69.90 & -- \\
FaceXRay~\cite{FaceXRay} & CVPR’20  & -- & 79.50 & -- & 80.58 & 80.92 & -- \\
F3Net~\cite{F3Net}& ECCV’20  & 98.20 & 78.88 & 71.77 & 81.11 & 73.50 & 76.32 \\
SPSL~\cite{SPSL} & CVPR’21  & 96.91 & 79.86 & 66.16 & 85.02 & 75.86 & 76.73 \\
RECCE~\cite{RECCE}  & CVPR’22  & 99.32 & 82.31 & 69.58 & 81.49 & 71.49 & 76.21 \\
SBI~\cite{SBI}  & CVPR’22  & 99.15 & 93.82 & 74.47 & 93.44 & 90.95 & 88.17 \\
UIA‑ViT~\cite{UIA-ViT}    & ECCV’22  & 99.33 & 82.41 & -- & 86.59 & 75.80 & -- \\
TALL~\cite{tall}        & ICCV’23  & \textbf{99.87} & 90.79 & 76.78 & -- & -- & -- \\
SeeABLE~\cite{SeeABLE}        & ICCV’23  & -- & 87.3 & 75.9 & -- & 86.3 & -- \\
AltFreezing~\cite{altfreezing}    & CVPR’23  & 93.81 & 89.50 & 64.75 & 88.48 & 64.05 & 76.70 \\
UCF~\cite{ucf}           & CVPR’23  & 98.69 & 83.73 & 75.11 & 86.08 & 80.50 & 81.36 \\
LSDA~\cite{lsda}          & CVPR’24  & -- & 91.10 & 77.00 & -- & -- & -- \\
CFM~\cite{CFM}           & TIFS’24  & -- & 89.65 & 80.22 & -- & -- & -- \\
InfoClue~\cite{InfoClue} & AAAI’24 & -- & 93.6 & 75.4 & -- & 90.2 & -- \\
LVLM‑DFD~\cite{LVLM-DFD}  & ICML’25 & 99.53 & \underline{94.71} & 79.12 & \textbf{97.62} & \textbf{91.81} & \underline{90.82} \\
VB‑StA~\cite{VB-StA}      & CVPR’25 & -- & 94.7 & \underline{84.3} & -- & 90.9 & -- \\
ProDet~\cite{prodet}      & NIPS’25  & -- & 92.62 & 71.52 & 94.48 & 82.83 & 85.36 \\
UDD~\cite{UDD}            & AAAI’25  & -- & 93.13 & 81.21 & -- & 88.11 & -- \\
\midrule
Ours & -- & \underline{99.56} & \textbf{95.26} & \textbf{89.96} & \underline{96.14} & \underline{91.57} & \textbf{93.23} \\
\bottomrule
\end{tabular}
\normalsize
\caption{ \textbf{Comparison of intra-dataset and cross-dataset performance between our method and existing deepfake detection methods.} The best AUC scores are highlighted in bold, and the second-best scores are underlined. All results are taken from the original publications or LVLM-DFD\cite{LVLM-DFD}.}
\label{tab:Cross-Dataset Evaluation}
\end{table*}

\section{Experiments}
\subsection{Experimental Settings}
\paragraph{Datasets} The FaceForensics++ (FF++)~\cite{ff++} dataset includes 1,000 real videos and 4,000 forgery videos across four deepfake categories, which is one of the most widely-used datasets for deepfake detection. CDF-v1, CDF-v2~\cite{celeb}, DFDCP~\cite{dfdcp}, DFDC~\cite{dfdc}, and DF40~\cite{df40} are commonly used datasets for evaluating generalization performance in deepfake detection. The datasets employed in this work are collected from DeepFakeBench~\cite{dfb} and DF40, which serve as standardized benchmarks for deepfake detection. To be consistent with the previous deepfake detection approaches, we trained only on c23 compression version of FF++ dataset.

\paragraph{Evaluation metrics} Following existing approaches, we report video-level Area Under the Receiver Operating Characteristic Curve (AUC) for a fair comparison with prior works. The video-level scores are computed by averaging the frame-level predictions across all frames in each video. 

\paragraph{Implementation Details}
We adopt the facebook/dinov2-with-registers-large~\cite{DINOv2} checkpoint as a frozen backbone and insert DFF-Adapter into the query, value, and dense projections of Transformer blocks. Each DFF-Adapter is configured with rank \( r = 16 \), scaling factor \( \alpha = 32 \), a total of 6 LoRA experts, and 4 head adapters.All the images are cropped to \(224 \times 224\) and the batch size is configured as 24.
We train the model for 50 epochs on a single NVIDIA RTX 4090 GPU using the Adam optimizer with a learning rate of \(2 \times 10^{-4}\) and a weight decay of \(1 \times 10^{-5}\). The loss weights in Eq.~(10) are set to \( \lambda_0 = 10 \) and \( \lambda_1 = 2 \). All experiments follow the default DeepfakeBench settings.

\subsection{Comparison with SOTA Detection Methods}
We compare our approach with several state-of-the-art deepfake detection methods.
\paragraph{Intra-Dataset Evaluation.} Following the intra-dataset evaluation protocol proposed in DeepfakeBench, we conduct a comprehensive comparison between our method and existing state-of-the-art deepfake detection approaches on the FF++ dataset. To ensure fairness and consistency, we strictly adhere to the training and testing splits defined in the benchmark. As reported in Table~\ref{tab:Cross-Dataset Evaluation} our method achieves a detection AUC of 99.56, demonstrating highly competitive performance and validating its effectiveness under standard evaluation settings.

\paragraph{Cross-Dataset Evaluation.} 
To assess the generalization ability of our method, we perform cross-dataset evaluations following the standardized protocol defined in DeepfakeBench. Specifically, the model is trained on the FF++ dataset (c23 version) and evaluated on several unseen datasets, including CDF-v1, CDF-v2, DFDCP, and DFDC. We report video-level AUC scores to comprehensively measure performance across diverse domains. As shown in Table~\ref{tab:Cross-Dataset Evaluation}, our method consistently outperforms existing state-of-the-art approaches on multiple challenging benchmarks, achieving an average improvement of 2.41 AUC points.
\begin{table}[t]
  \centering
  \small
  \renewcommand\arraystretch{1.15}
  \setlength{\tabcolsep}{2pt}
  \begin{tabular}{@{}ccccccc@{}}
    \toprule
    \multirow{5}{*}{\rotatebox{90}{DF40-FS}} &  &  & FaceDancer & InSwapper & FSGAN & UniFace \\ \cmidrule(l){3-7}
    &  & SBI      & 78.18 & 88.52 & 89.62 & 89.02 \\
    &  & DINOv2   & 57.31 & 64.48 & 77.24 & 68.76 \\
    &  & LVLM-DFD & 82.97 & 87.64 & 93.75 & 90.61 \\
    &  & Ours     & \textbf{93.15} & \textbf{93.98} & \textbf{98.84} & \textbf{94.51} \\ 
    \midrule\midrule
    \multirow{5}{*}{\rotatebox{90}{DF40-FR}} &  &  & FOMM & HyperReenact & Wav2Lip & MCNet \\ \cmidrule(l){3-7}
    &  & SBI      & 88.03 & 65.26 & 77.06 & 81.51 \\
    &  & DINOv2   & 76.67 & 46.25 & 56.36 & 57.40 \\
    &  & LVLM-DFD & 93.34 & 81.56 & 78.60 & 83.45 \\
    &  & Ours     & \textbf{95.36} & \textbf{90.28} & \textbf{91.65} & \textbf{86.29} \\
    \midrule\midrule
    \multirow{5}{*}{\rotatebox{90}{DF40-EFS}} &  &  & StyleGAN3 & StyleGAN-XL & VQGAN & DiT-XL/2 \\ \cmidrule(l){3-7}
    &  & SBI      & 97.91 & 23.26 & 91.47 & 53.59 \\
    &  & DINOv2   & 85.93 & 87.70 & 90.21 & 74.57 \\
    &  & LVLM-DFD & 98.87 & 100.00 & 99.99 & 86.61 \\
    &  & Ours     & \textbf{99.92} & \textbf{100.00} & \textbf{100.00} & \textbf{98.44} \\
    \bottomrule
  \end{tabular}
  \normalsize
  \caption{\textbf{Cross-manipulation evaluation on DF40.} All methods are evaluated on different manipulated subsets. Best results are in bold.}
  \label{tb:cross-manipulation}
\end{table}

\paragraph{Cross-Manipulation Evaluation.}
With the rapid advancement of generative technologies, new forgery techniques continue to emerge at an unprecedented pace. As a result, many existing detection models exhibit suboptimal performance on the recently introduced DF40 dataset.  To evaluate the generalization capability of our method against unseen forgery techniques, we train the model on the FF++ dataset and conduct testing on the DF40 benchmark. We evaluate performance across the FF++ domain of DF40, covering three major forgery categories: Face-swapping (FS), Face-reenactment (FR), and Entire Face Synthesis(EFS), spanning a total of 12 diverse manipulation methods.  As presented in Table~\ref{tb:cross-manipulation}, our method outperforms all existing detection approaches across all categories, demonstrating strong generalization ability in detecting a wide range of unseen forgery techniques.

\subsection{Analysis}
\paragraph{Ablation studies.} 
To evaluate the effectiveness of the proposed Forgery-Aware Multi-Head Router (FAMHR) and Shared-Enhanced Task Fusion (SETF), we perform ablation studies across multiple datasets. As shown in Table~\ref{tab:ablation}, introducing FAMHR significantly improves cross-dataset performance by enabling adaptive expert routing over feature subspaces, which enhances the model’s ability to capture diverse localized forgery artefacts. Building on this, incorporating SETF consistently boosts performance by transferring fine-grained cues from the auxiliary task to the main authenticity stream. The full model achieves the highest AUC across all benchmarks, demonstrating the complementary benefits of FAMHR and SETF in enhancing generalization under diverse manipulation types.More ablation results are provided in the Appendix.
\paragraph{Comparison with Fine-Tuning Strategies.} 
We conduct a comprehensive evaluation of the proposed DFF-Adapter for deepfake detection by comparing it with several representative fine tuning strategies : (1) Linear Probing (LP), which freezes the backbone and trains only a linear classifier; (2) LoRA fine-tuning, which inserts low-rank adaptation matrices for efficient updates; (3) MoE-FFD~\cite{MoE-FFD}, a recent approach that integrates LoRA and Adapter layers with a mixture-of-experts routing mechanism for forgery detection; and (4) our method. As reported in Table~\ref{tab:reprogramming paradigm}, DFF-Adapter achieves superior performance across diverse experimental settings, demonstrating its effectiveness in capturing manipulation-specific features and strong generalization ability to unseen forgeries.This superior performance is attributed to its ability to fully exploit DINOv2’s strength in local feature representation, enabling the model to capture fine-grained forgery artifacts more effectively.
\begin{table}[t]
\centering
\small
\setlength{\tabcolsep}{3pt} 
\begin{tabular}{ccc|cccc}
\toprule
DINOv2&FAMRH & SETF & 
CDF-v2 & DFDC & 
CDF-v1 & DFDCP \\
\midrule
\cmark & &  & 61.93 & 52.83 & 68.46 & 55.54 \\
\cmark &\cmark &  & 89.56 & 86.24 & 85.53 & 87.11 \\
\cmark &\cmark & \cmark & \textbf{95.26} & \textbf{89.96} & \textbf{96.14} & \textbf{91.57} \\
\bottomrule
\end{tabular}
\normalsize
\caption{\textbf{Ablation study results on cross-dataset evaluation.}}
\label{tab:ablation}
\end{table}

\begin{table}[t]
\centering
\small
\setlength{\tabcolsep}{4pt} 
\begin{tabular}{c|cccc}
\toprule
Fine-Tuning Paradigms & 
CDF-v2 & DFDC & 
CDF-v1 & DFDCP \\
\midrule
LP  &66.63 & 72.54 & 56.45 & 65.90 \\
LoRa  & 85.14 & 79.95 & 85.14 & 81.49 \\
MoE-FFD & 80.40 & 76.52 & 65.75 & 87.60 \\
ours & \textbf{95.26} & \textbf{89.96} & \textbf{96.14} & \textbf{91.57} \\
\bottomrule
\end{tabular}
\normalsize
\caption{\textbf{Comparison with fine-tuning strategies on cross-dataset evaluation. }}
\label{tab:reprogramming paradigm}
\end{table}

\paragraph{Identity Constrained Training.} As concerns over biometric privacy continue to rise, it is becoming increasingly difficult to obtain large-scale face datasets, particularly in terms of acquiring a sufficient number of identities and training images. To rigorously evaluate the generalization capability of our proposed DFF-Adapter-DINO detector under such data-scarce conditions, we conducted a series of experiments with progressively fewer training identities. Specifically, we randomly sampled 10, 30, and 50 identities from the FF++ training set, which correspond to approximately 1\%, 3\%, and 5\% of the total available identities, respectively. As shown in Table~\ref{tab:Identity-Constrained}, our method achieves decent performance with only 50 identities. Even with as few as 10  identities, it maintains competitive accuracy. These results demonstrate the generalization strength and practical relevance of our method in few-identity settings, indicating that DFF-Adapter successfully harnesses the visual priors of DINOv2 for generalizable forgery detection.
\begin{table}[t]
\centering
\small
\setlength{\tabcolsep}{4pt} 
\begin{tabular}{c|cccc}
\toprule
Training Identities & 
CDF-v2 & DFDC & 
CDF-v1 & DFDCP \\
\midrule
10  &81.97 & \textbf{82.63} & 81.62 & 82.22 \\
30  & 81.94 & 79.16 & 90.49 & 81.25 \\
50 & \textbf{87.28} & 82.37 & \textbf{91.26} & \textbf{86.16} \\
\bottomrule
\end{tabular}
\normalsize
\caption{\textbf{Cross-dataset evaluation under identity constrained training.}}
\label{tab:Identity-Constrained}
\end{table}

\begin{table}[t]
  \centering
  \small
  {
  \setlength{\tabcolsep}{3pt}
  \renewcommand{\arraystretch}{1.05}
  \begin{tabular}{@{}l c cc cc cc @{}}
    \toprule
    \multirow{2}{*}{Model} & \multirow{2}{*}{Arch.} &
    \multicolumn{2}{c}{FF++} &
    \multicolumn{2}{c}{CDF-v2} &
    \multicolumn{2}{c}{DFDC} \\
    \cmidrule(lr){3-4} \cmidrule(lr){5-6} \cmidrule(lr){7-8}
    & & AUC & EER & AUC & EER & AUC & EER \\
    \midrule
    CLIP$^{\phantom{*}}$     & ViT-B/16 & 86.01 & 22.85 & 76.27 & 30.89 & 77.20 & 30.06 \\
    CLIP$^{\phantom{*}}$     & ViT-L/14 & 91.30 & 17.14 & 80.16 & 29.17 & 76.76 & 30.46 \\
    CLIP$^{*}$               & ViT-B/16 & 99.29 & \textbf{0.71} & 85.19 & 22.47 & 83.19 & 24.82 \\
    CLIP$^{*}$               & ViT-L/14 & 97.14 & 2.57 & 89.10 & 15.28 & 84.75 & 23.32 \\
    DINOv2$^{\phantom{*}}$   & ViT-B/16 & 79.07 & 28.57 & 65.38 & 40.45 & 68.97 & 36.58 \\
    DINOv2$^{\phantom{*}}$   & ViT-L/14 & 84.84 & 23.57 & 66.63 & 35.39 & 72.54 & 33.15 \\
    DINOv2$^{*}$             & ViT-B/16 & \textbf{99.61} & 2.32 & 93.35 & 14.61 & 82.00 & 25.49 \\
    DINOv2$^{*}$             & ViT-L/14 & 99.56 & 1.79 & \textbf{95.26} & \textbf{12.65} & \textbf{89.96} & \textbf{17.96} \\
    \bottomrule
  \end{tabular}
  }
  \normalsize
  \caption{\textbf{Comparison with pre-trained large models on FF++, CDF-v2, and DFDC.} Performance is reported as AUC (\%) and EER (\%).}
  \label{tab:LLM}
\end{table}

\paragraph{Comparison with Pre-trained Large Models.}
As shown in Table~\ref{tab:LLM}, we conducted a comparative study of representative pre-trained vision models, including CLIP and DINO, and evaluated their performance using AUC and Equal Error Rate (EER). Models marked with * denote those integrated with DFF-Adapter. CLIP ,trained on image–text pairs, emphasizes semantic alignment but lacks sensitivity to fine-grained visual artifacts. DINO improves upon CLIP by capturing local structures through self-distillation. To fully leverage the rich visual priors encoded in DINOv2, we propose DFF-Adapter. Our method jointly optimizes authenticity prediction and forgery-type classification, enabling more discriminative and manipulation-sensitive representations through multi-head routing and shared knowledge transfer. Built on this synergistic design, our method achieves the highest AUCs and lowest EERs across all evaluated benchmarks. 

\paragraph{t-SNE Feature Visualization.}
To illustrate the effectiveness of DFF-Adapter in learning discriminative and generalizable representations, we conduct a t-SNE visualization of the learned feature distributions under both intra-dataset and cross-dataset settings. Specifically, we compare the features extracted from a vanilla DINOv2 model (fine-tuned only at the last layer) and our full DFF-Adapter model. The visualization is performed on the FF++ dataset (intra-dataset) and the unseen CDF-v2 dataset (cross-dataset). 

\begin{figure}[t]
  \centering
  \begin{minipage}{0.48\linewidth}
    \centering
    \includegraphics[width=\linewidth]{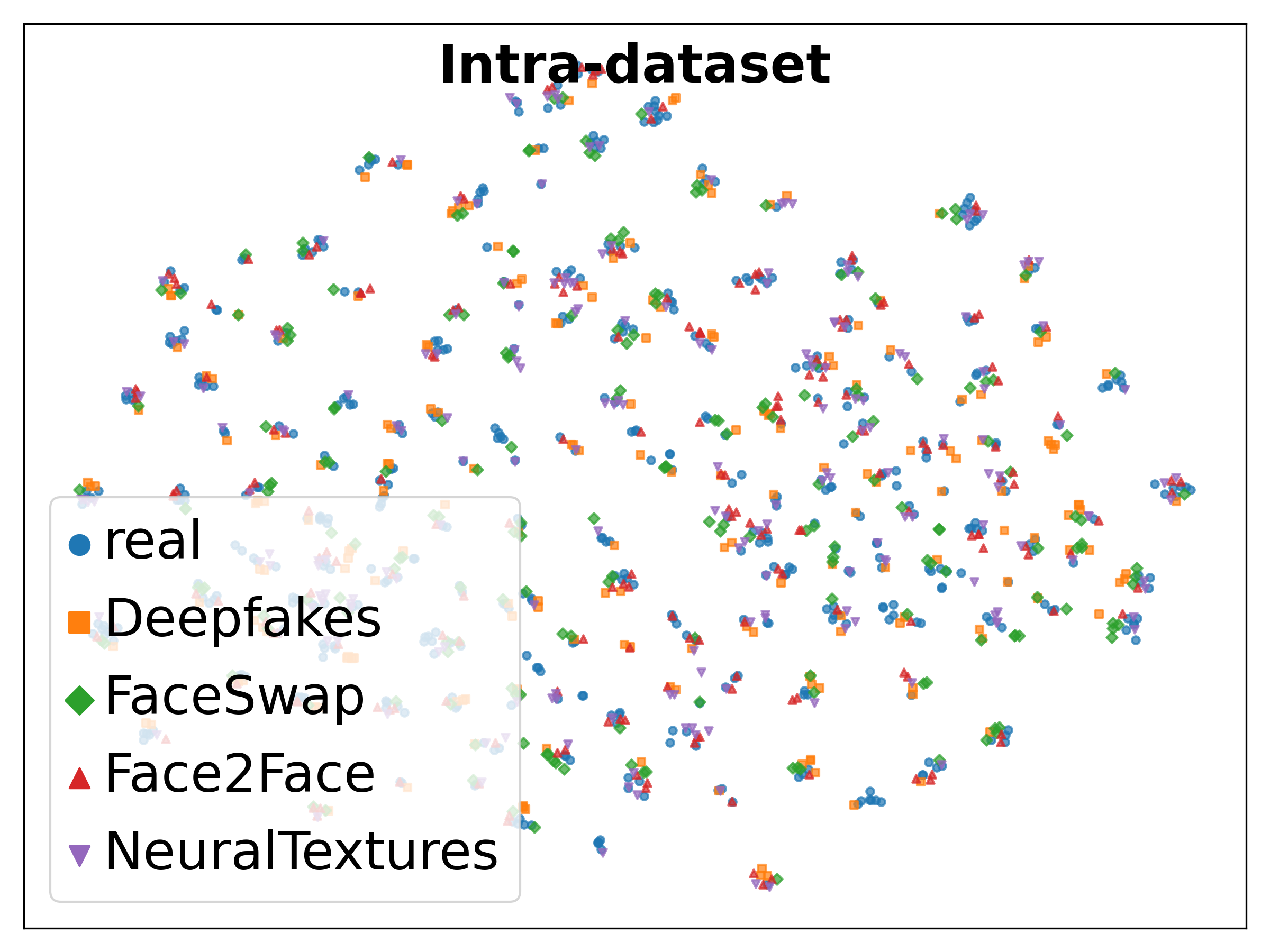}\\[2pt]
    \includegraphics[width=\linewidth]{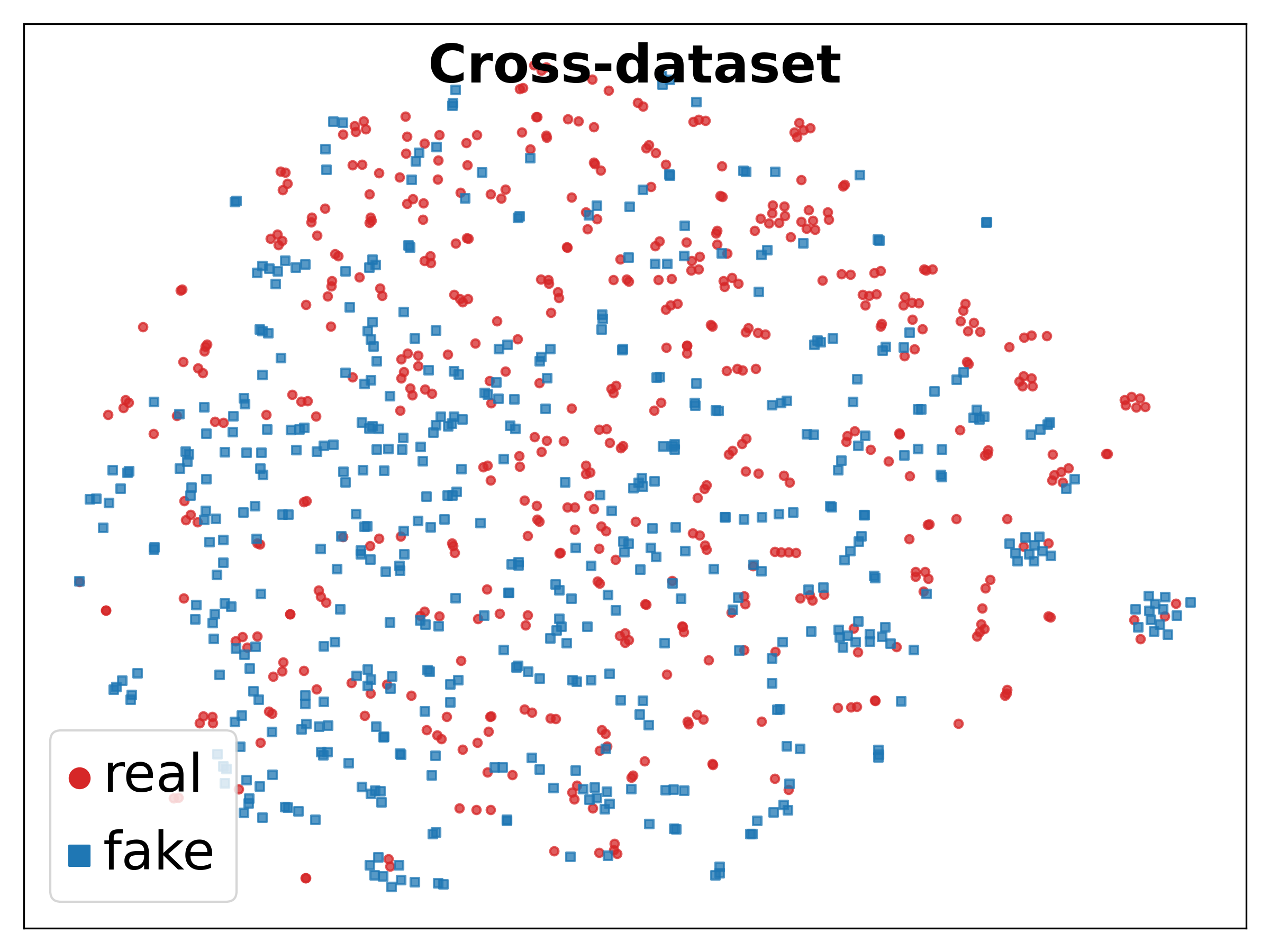}\\[2pt]
    \textbf{(a) DINOv2}
  \end{minipage}
  \hfill
  \begin{minipage}{0.48\linewidth}
    \centering
    \includegraphics[width=\linewidth]{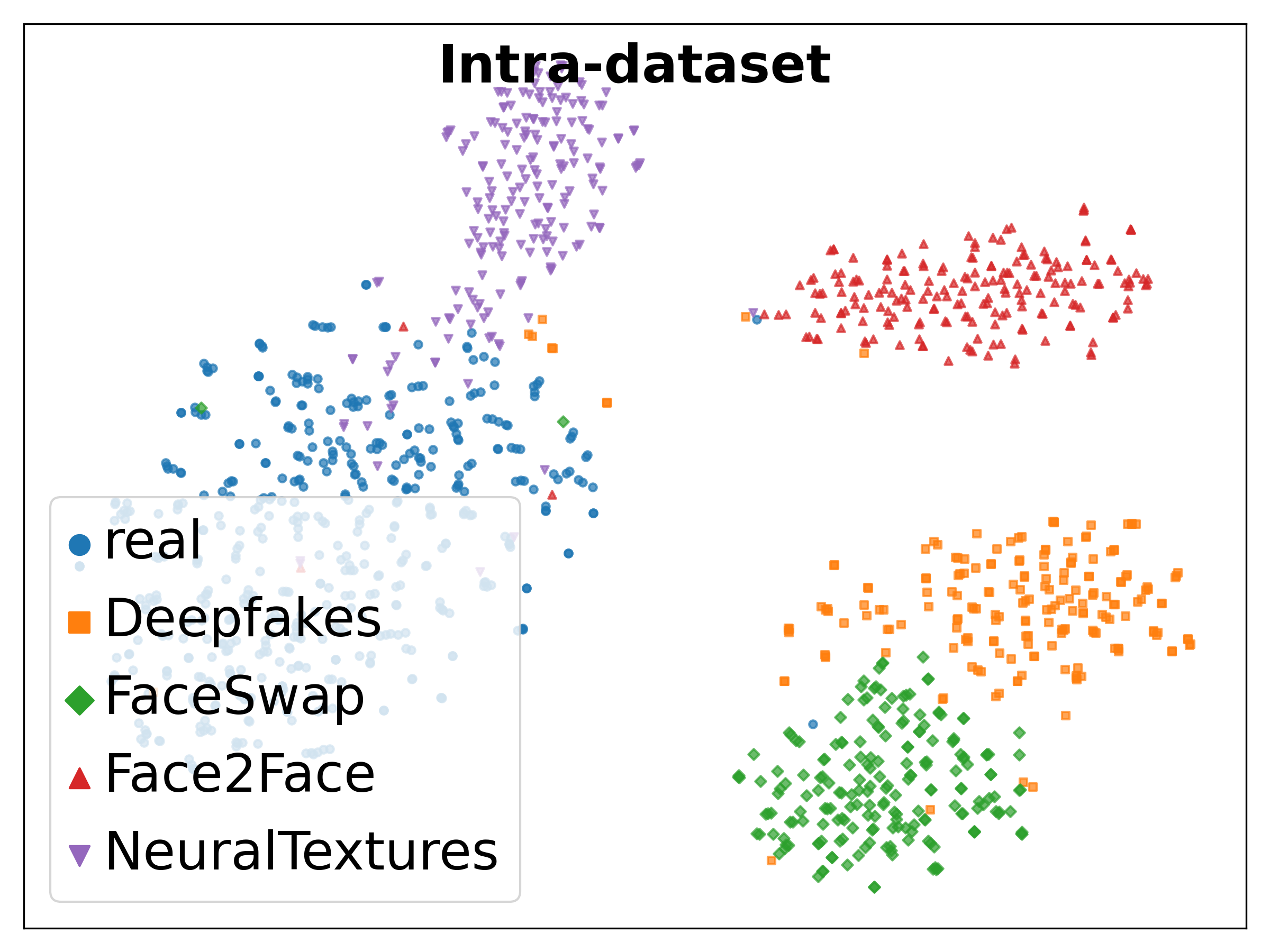}\\[2pt]
    \includegraphics[width=\linewidth]{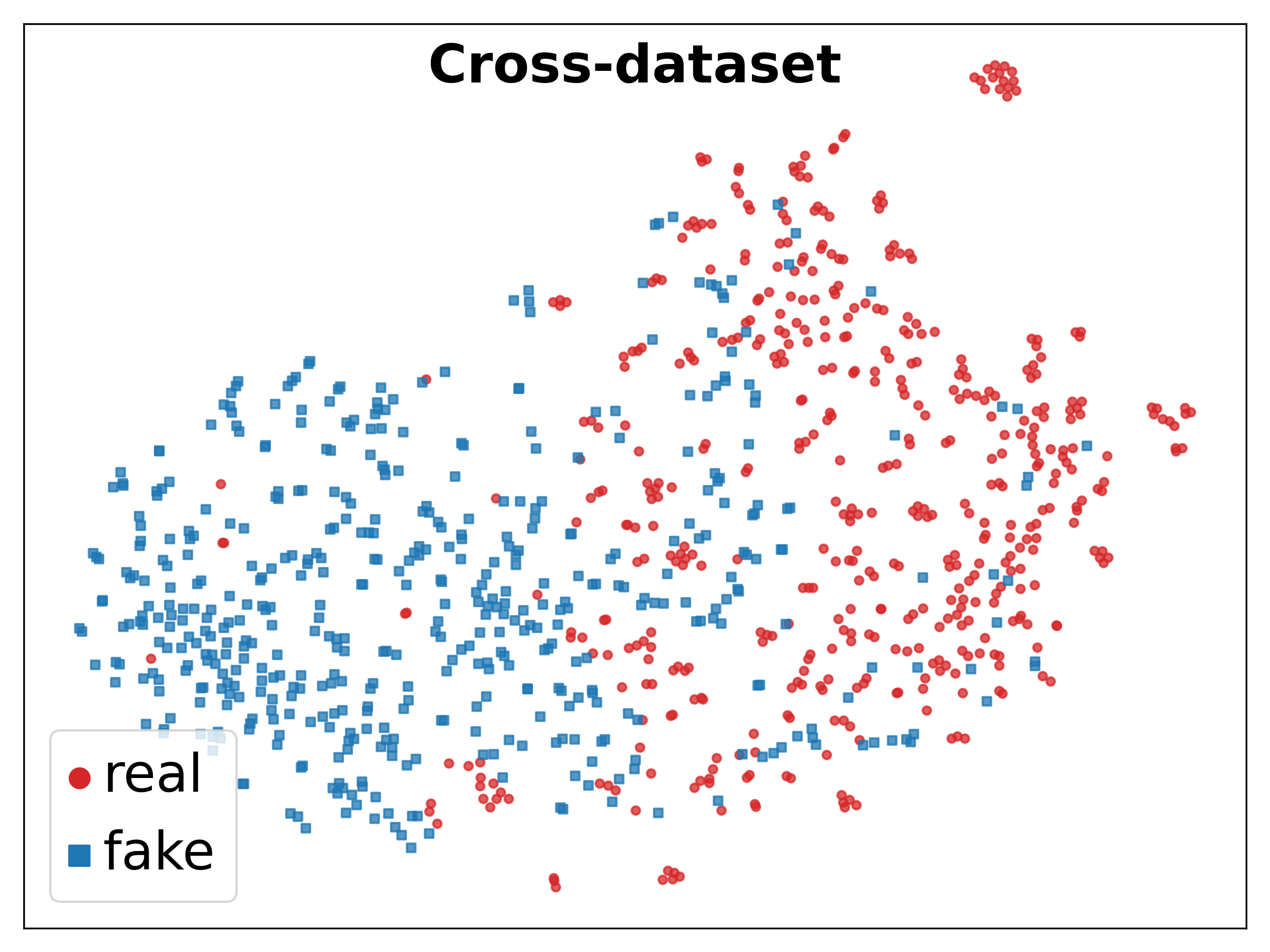}\\[2pt]
    \textbf{(b) OURS}
  \end{minipage}

  \caption{T\textendash{}SNE visualizations on two datasets: top row shows intra-dataset (FF++), bottom row shows cross-dataset (CDF-v2).}
  \label{fig:tsne}
\end{figure}

As shown in Figure.~\ref{fig:tsne}, in the intra-dataset setting (FF++), the DINOv2 baseline fails to distinguish between real and fake samples, as its t-SNE plot exhibits a highly entangled distribution with no clear separation between authentic and manipulated faces. This indicates that simply fine-tuning the final layer is insufficient to extract meaningful features for forgery detection. In contrast, our DFF-Adapter yields well-separated and compact clusters in the feature space—not only achieving clear boundaries between real and fake samples, but also effectively separating different types of forgeries. This demonstrates that the auxiliary forgery-type classification task enables the authenticity branch to benefit from manipulation-specific cues, resulting in more discriminative and semantically structured feature representations.In the cross-dataset visualization on CDF-v2, the difference becomes even more pronounced. The baseline DINOv2 model shows entangled and irregular feature clusters, highlighting its poor generalization to out-of-distribution samples. Meanwhile, our DFF-Adapter maintains clear separation between real and fake samples, and still preserves identifiable structure among different forgery types.

\subsection{Conclusion}
In this work, we introduced DFF-Adapter-DINO, which enriches a frozen DINOv2 backbone with DeepFake Fine-Grained Adapters. By jointly optimizing authenticity detection and forgery-type classification, and by sharing fine-grained cues through a dedicated shared adapter, DFF-Adapter-DINO leverages both global semantics and local artifact patterns.Across five deepfake benchmarks, our method attains the best overall performance and maintains strong generalization under challenging cross-dataset and cross-manipulation scenarios. Taken together, these results demonstrate that DFF-Adapter provides an efficient way to leverage the rich priors of DINOv2 for face forgery detection.

In future work, we will enhance the detection performance of our method on forged data generated by the latest models and design efficient algorithms to tackle real-world scenarios.

\section{Acknowledgments}
This work is supported in part by the National Natural Science Foundation of China under grant numbers, U23B2023 and 62472199, Guangdong Key Laboratory of Data Security and Privacy Preserving under Grant 2023B1212060036, the basic and Applied Basic Research Foundation of Guangdong Province (2025A1515011097), and the Outstanding Youth Project of Guangdong Basic and Applied Basic Research Foundation (2023B1515020064). This work is also supported by Engineering Research Center of Trustworthy AI, Ministry of Education.)

\bibliography{aaai2026}

@inproceedings{ILLUSION,
  title={ILLUSION: Unveiling truth with a comprehensive multi-modal, multi-lingual deepfake dataset},
  author={Thakral, Kartik and Ranjan, Rishabh and Singh, Akanksha and Jain, Akshat and Vatsa, Mayank and Singh, Richa},
  booktitle={The Thirteenth International Conference on Learning Representations},
  year={2025}
}

@inproceedings{AI-Face,
  title={Ai-face: A million-scale demographically annotated ai-generated face dataset and fairness benchmark},
  author={Lin, Li and Santosh, Santosh and Wu, Mingyang and Wang, Xin and Hu, Shu},
  booktitle={Proceedings of the Computer Vision and Pattern Recognition Conference},
  pages={3503--3515},
  year={2025}
}

@inproceedings{lips,
  title={Lips don't lie: A generalisable and robust approach to face forgery detection},
  author={Haliassos, Alexandros and Vougioukas, Konstantinos and Petridis, Stavros and Pantic, Maja},
  booktitle={Proceedings of the IEEE/CVF conference on computer vision and pattern recognition},
  pages={5039--5049},
  year={2021}
}

@inproceedings{frepgan,
  title={Frepgan: robust deepfake detection using frequency-level perturbations},
  author={Jeong, Yonghyun and Kim, Doyeon and Ro, Youngmin and Choi, Jongwon},
  booktitle={Proceedings of the AAAI conference on artificial intelligence},
  volume={36},
  pages={1060--1068},
  year={2022}
}

@inproceedings{dynamic,
  title={Dynamic graph learning with content-guided spatial-frequency relation reasoning for deepfake detection},
  author={Wang, Yuan and Yu, Kun and Chen, Chen and Hu, Xiyuan and Peng, Silong},
  booktitle={Proceedings of the IEEE/CVF conference on computer vision and pattern recognition},
  pages={7278--7287},
  year={2023}
}

@inproceedings{frequency,
  title={Frequency-aware deepfake detection: Improving generalizability through frequency space domain learning},
  author={Tan, Chuangchuang and Zhao, Yao and Wei, Shikui and Gu, Guanghua and Liu, Ping and Wei, Yunchao},
  booktitle={Proceedings of the AAAI Conference on Artificial Intelligence},
  volume={38},
  pages={5052--5060},
  year={2024}
}

@inproceedings{tall,
  title={Tall: Thumbnail layout for deepfake video detection},
  author={Xu, Yuting and Liang, Jian and Jia, Gengyun and Yang, Ziming and Zhang, Yanhao and He, Ran},
  booktitle={Proceedings of the IEEE/CVF international conference on computer vision},
  pages={22658--22668},
  year={2023}
}

@inproceedings{uia,
  title={UIA-ViT: Unsupervised inconsistency-aware method based on vision transformer for face forgery detection},
  author={Zhuang, Wanyi and Chu, Qi and Tan, Zhentao and Liu, Qiankun and Yuan, Haojie and Miao, Changtao and Luo, Zixiang and Yu, Nenghai},
  booktitle={European conference on computer vision},
  pages={391--407},
  year={2022},
  organization={Springer}
}

@article{survey1,
  title={Deepfakes and beyond: A survey of face manipulation and fake detection},
  author={Tolosana, Ruben and Vera-Rodriguez, Ruben and Fierrez, Julian and Morales, Aythami and Ortega-Garcia, Javier},
  journal={Information Fusion},
  volume={64},
  pages={131--148},
  year={2020},
  publisher={Elsevier}
}

@article{survey2,
  title={Deepfakes generation and detection: State-of-the-art, open challenges, countermeasures, and way forward},
  author={Masood, Momina and Nawaz, Mariam and Malik, Khalid Mahmood and Javed, Ali and Irtaza, Aun and Malik, Hafiz},
  journal={Applied intelligence},
  volume={53},
  number={4},
  pages={3974--4026},
  year={2023},
  publisher={Springer}
}

@article{survey3,
  title={Deepfake generation and detection: A benchmark and survey},
  author={Pei, Gan and Zhang, Jiangning and Hu, Menghan and Zhang, Zhenyu and Wang, Chengjie and Wu, Yunsheng and Zhai, Guangtao and Yang, Jian and Shen, Chunhua and Tao, Dacheng},
  journal={arXiv preprint arXiv:2403.17881},
  year={2024}
}

@article{DINOv2,
  title={Dinov2: Learning robust visual features without supervision},
  author={Oquab, Maxime and Darcet, Timoth{\'e}e and Moutakanni, Th{\'e}o and Vo, Huy and Szafraniec, Marc and Khalidov, Vasil and Fernandez, Pierre and Haziza, Daniel and Massa, Francisco and El-Nouby, Alaaeldin and others},
  journal={arXiv preprint arXiv:2304.07193},
  year={2023}
}

@inproceedings{expss,
  title={Exploring self-supervised vision transformers for deepfake detection: A comparative analysis},
  author={Nguyen, Huy H and Yamagishi, Junichi and Echizen, Isao},
  booktitle={2024 IEEE International Joint Conference on Biometrics (IJCB)},
  pages={1--10},
  year={2024},
  organization={IEEE}
}

@inproceedings{pudd,
  title={Pudd: Towards robust multi-modal prototype-based deepfake detection},
  author={Pellicer, Alvaro Lopez and Li, Yi and Angelov, Plamen},
  booktitle={Proceedings of the IEEE/CVF Conference on Computer Vision and Pattern Recognition},
  pages={3809--3817},
  year={2024}
}

@inproceedings{dino,
  title={Emerging properties in self-supervised vision transformers},
  author={Caron, Mathilde and Touvron, Hugo and Misra, Ishan and J{\'e}gou, Herv{\'e} and Mairal, Julien and Bojanowski, Piotr and Joulin, Armand},
  booktitle={Proceedings of the IEEE/CVF international conference on computer vision},
  pages={9650--9660},
  year={2021}
}

@inproceedings{exploring,
  title={Exploring unbiased deepfake detection via token-level shuffling and mixing},
  author={Fu, Xinghe and Yan, Zhiyuan and Yao, Taiping and Chen, Shen and Li, Xi},
  booktitle={Proceedings of the AAAI Conference on Artificial Intelligence},
  volume={39},
  pages={3040--3048},
  year={2025}
}

@article{truthlens,
  title={TruthLens: Explainable DeepFake Detection for Face Manipulated and Fully Synthetic Data},
  author={Kundu, Rohit and Balachandran, Athula and Roy-Chowdhury, Amit K},
  journal={arXiv preprint arXiv:2503.15867},
  year={2025}
}

@inproceedings{ff++,
  title={Faceforensics++: Learning to detect manipulated facial images},
  author={Rossler, Andreas and Cozzolino, Davide and Verdoliva, Luisa and Riess, Christian and Thies, Justus and Nie{\ss}ner, Matthias},
  booktitle={Proceedings of the IEEE/CVF international conference on computer vision},
  pages={1--11},
  year={2019}
}

@inproceedings{celeb,
  title={Celeb-df: A large-scale challenging dataset for deepfake forensics},
  author={Li, Yuezun and Yang, Xin and Sun, Pu and Qi, Honggang and Lyu, Siwei},
  booktitle={Proceedings of the IEEE/CVF conference on computer vision and pattern recognition},
  pages={3207--3216},
  year={2020}
}

@article{dfdc,
  title={The deepfake detection challenge (dfdc) dataset},
  author={Dolhansky, Brian and Bitton, Joanna and Pflaum, Ben and Lu, Jikuo and Howes, Russ and Wang, Menglin and Ferrer, Cristian Canton},
  journal={arXiv preprint arXiv:2006.07397},
  year={2020}
}

@article{dfdcp,
  title={The deepfake detection challenge (dfdc) dataset},
  author={Dolhansky, Brian and Bitton, Joanna and Pflaum, Ben and Lu, Jikuo and Howes, Russ and Wang, Menglin and Ferrer, Cristian Canton},
  journal={arXiv preprint arXiv:2006.07397},
  year={2020}
}

@article{df40,
  title={Df40: Toward next-generation deepfake detection},
  author={Yan, Zhiyuan and Yao, Taiping and Chen, Shen and Zhao, Yandan and Fu, Xinghe and Zhu, Junwei and Luo, Donghao and Wang, Chengjie and Ding, Shouhong and Wu, Yunsheng and others},
  journal={Advances in Neural Information Processing Systems},
  volume={37},
  pages={29387--29434},
  year={2024}
}

@article{fwa,
  title={Exposing deepfake videos by detecting face warping artifacts},
  author={Li, Yuezun and Lyu, Siwei},
  journal={arXiv preprint arXiv:1811.00656},
  year={2018}
}

@article{delocate,
  title={Delocate: Detection and localization for deepfake videos with randomly-located tampered traces},
  author={Hu, Juan and Liao, Xin and Gao, Difei and Tsutsui, Satoshi and Wang, Qian and Qin, Zheng and Shou, Mike Zheng},
  journal={arXiv preprint arXiv:2401.13516},
  year={2024}
}

@inproceedings{lec,
  title={Learning self-consistency for deepfake detection},
  author={Zhao, Tianchen and Xu, Xiang and Xu, Mingze and Ding, Hui and Xiong, Yuanjun and Xia, Wei},
  booktitle={Proceedings of the IEEE/CVF international conference on computer vision},
  pages={15023--15033},
  year={2021}
}

@inproceedings{wsc,
  title={Towards generic image manipulation detection with weakly-supervised self-consistency learning},
  author={Zhai, Yuanhao and Luan, Tianyu and Doermann, David and Yuan, Junsong},
  booktitle={Proceedings of the IEEE/CVF International Conference on Computer Vision},
  pages={22390--22400},
  year={2023}
}

@inproceedings{VB-StA,
  title={Generalizing deepfake video detection with plug-and-play: Video-level blending and spatiotemporal adapter tuning},
  author={Yan, Zhiyuan and Zhao, Yandan and Chen, Shen and Guo, Mingyi and Fu, Xinghe and Yao, Taiping and Ding, Shouhong and Wu, Yunsheng and Yuan, Li},
  booktitle={Proceedings of the Computer Vision and Pattern Recognition Conference},
  pages={12615--12625},
  year={2025}
}

@inproceedings{RepDFD,
  title={Standing on the shoulders of giants: Reprogramming visual-language model for general deepfake detection},
  author={Lin, Kaiqing and Lin, Yuzhen and Li, Weixiang and Yao, Taiping and Li, Bin},
  booktitle={Proceedings of the AAAI Conference on Artificial Intelligence},
  volume={39},
  pages={5262--5270},
  year={2025}
}

@article{LVLM-DFD,
  title={Unlocking the Capabilities of Large Vision-Language Models for Generalizable and Explainable Deepfake Detection},
  author={Yu, Peipeng and Fei, Jianwei and Gao, Hui and Feng, Xuan and Xia, Zhihua and Chang, Chip Hong},
  journal={arXiv preprint arXiv:2503.14853},
  year={2025}
}

@inproceedings{FaceXRay,
  title={Face x-ray for more general face forgery detection},
  author={Li, Lingzhi and Bao, Jianmin and Zhang, Ting and Yang, Hao and Chen, Dong and Wen, Fang and Guo, Baining},
  booktitle={Proceedings of the IEEE/CVF conference on computer vision and pattern recognition},
  pages={5001--5010},
  year={2020}
}

@inproceedings{UIA-ViT,
  title={UIA-ViT: Unsupervised inconsistency-aware method based on vision transformer for face forgery detection},
  author={Zhuang, Wanyi and Chu, Qi and Tan, Zhentao and Liu, Qiankun and Yuan, Haojie and Miao, Changtao and Luo, Zixiang and Yu, Nenghai},
  booktitle={European conference on computer vision},
  pages={391--407},
  year={2022},
  organization={Springer}
}

@inproceedings{altfreezing,
  title={Altfreezing for more general video face forgery detection},
  author={Wang, Zhendong and Bao, Jianmin and Zhou, Wengang and Wang, Weilun and Li, Houqiang},
  booktitle={Proceedings of the IEEE/CVF conference on computer vision and pattern recognition},
  pages={4129--4138},
  year={2023}
}

@inproceedings{ucf,
  title={Ucf: Uncovering common features for generalizable deepfake detection},
  author={Yan, Zhiyuan and Zhang, Yong and Fan, Yanbo and Wu, Baoyuan},
  booktitle={Proceedings of the IEEE/CVF international conference on computer vision},
  pages={22412--22423},
  year={2023}
}

@inproceedings{lsda,
  title={Transcending forgery specificity with latent space augmentation for generalizable deepfake detection},
  author={Yan, Zhiyuan and Luo, Yuhao and Lyu, Siwei and Liu, Qingshan and Wu, Baoyuan},
  booktitle={Proceedings of the IEEE/CVF Conference on Computer Vision and Pattern Recognition},
  pages={8984--8994},
  year={2024}
}

@article{prodet,
  title={Can we leave deepfake data behind in training deepfake detector?},
  author={Cheng, Jikang and Yan, Zhiyuan and Zhang, Ying and Luo, Yuhao and Wang, Zhongyuan and Li, Chen},
  journal={Advances in Neural Information Processing Systems},
  volume={37},
  pages={21979--21998},
  year={2024}
}

@article{CFM,
  title={Beyond the prior forgery knowledge: Mining critical clues for general face forgery detection},
  author={Luo, Anwei and Kong, Chenqi and Huang, Jiwu and Hu, Yongjian and Kang, Xiangui and Kot, Alex C},
  journal={IEEE Transactions on Information Forensics and Security},
  volume={19},
  pages={1168--1182},
  year={2023},
  publisher={IEEE}
}

@inproceedings{UDD,
  title={Exploring unbiased deepfake detection via token-level shuffling and mixing},
  author={Fu, Xinghe and Yan, Zhiyuan and Yao, Taiping and Chen, Shen and Li, Xi},
  booktitle={Proceedings of the AAAI Conference on Artificial Intelligence},
  volume={39},
  pages={3040--3048},
  year={2025}
}

@inproceedings{InfoClue,
  title={Exposing the deception: Uncovering more forgery clues for deepfake detection},
  author={Ba, Zhongjie and Liu, Qingyu and Liu, Zhenguang and Wu, Shuang and Lin, Feng and Lu, Li and Ren, Kui},
  booktitle={Proceedings of the AAAI Conference on Artificial Intelligence},
  volume={38},
  pages={719--728},
  year={2024}
}

@article{dfb,
  title={Deepfakebench: A comprehensive benchmark of deepfake detection},
  author={Yan, Zhiyuan and Zhang, Yong and Yuan, Xinhang and Lyu, Siwei and Wu, Baoyuan},
  journal={arXiv preprint arXiv:2307.01426},
  year={2023}
}

@inproceedings{SPSL,
  title={Spatial-phase shallow learning: rethinking face forgery detection in frequency domain},
  author={Liu, Honggu and Li, Xiaodan and Zhou, Wenbo and Chen, Yuefeng and He, Yuan and Xue, Hui and Zhang, Weiming and Yu, Nenghai},
  booktitle={Proceedings of the IEEE/CVF conference on computer vision and pattern recognition},
  pages={772--781},
  year={2021}
}

@inproceedings{RECCE,
  title={End-to-end reconstruction-classification learning for face forgery detection},
  author={Cao, Junyi and Ma, Chao and Yao, Taiping and Chen, Shen and Ding, Shouhong and Yang, Xiaokang},
  booktitle={Proceedings of the IEEE/CVF conference on computer vision and pattern recognition},
  pages={4113--4122},
  year={2022}
}

@inproceedings{SBI,
  title={Detecting deepfakes with self-blended images},
  author={Shiohara, Kaede and Yamasaki, Toshihiko},
  booktitle={Proceedings of the IEEE/CVF conference on computer vision and pattern recognition},
  pages={18720--18729},
  year={2022}
}

@inproceedings{F3Net,
  title={Thinking in frequency: Face forgery detection by mining frequency-aware clues},
  author={Qian, Yuyang and Yin, Guojun and Sheng, Lu and Chen, Zixuan and Shao, Jing},
  booktitle={European conference on computer vision},
  pages={86--103},
  year={2020},
  organization={Springer}
}

@article{MoE-FFD,
  title={Moe-ffd: Mixture of experts for generalized and parameter-efficient face forgery detection},
  author={Kong, Chenqi and Luo, Anwei and Bao, Peijun and Yu, Yi and Li, Haoliang and Zheng, Zengwei and Wang, Shiqi and Kot, Alex C},
  journal={arXiv preprint arXiv:2404.08452},
  year={2024}
}

@article{LoRA,
  title={Lora: Low-rank adaptation of large language models.},
  author={Hu, Edward J and Shen, Yelong and Wallis, Phillip and Allen-Zhu, Zeyuan and Li, Yuanzhi and Wang, Shean and Wang, Lu and Chen, Weizhu and others},
  journal={ICLR},
  volume={1},
  number={2},
  pages={3},
  year={2022}
}

@inproceedings{SeeABLE,
  title={Seeable: Soft discrepancies and bounded contrastive learning for exposing deepfakes},
  author={Larue, Nicolas and Vu, Ngoc-Son and Struc, Vitomir and Peer, Peter and Christophides, Vassilis},
  booktitle={Proceedings of the IEEE/CVF International Conference on Computer Vision},
  pages={21011--21021},
  year={2023}
}

@inproceedings{finfer,
  title={Finfer: Frame inference-based deepfake detection for high-visual-quality videos},
  author={Hu, Juan and Liao, Xin and Liang, Jinwen and Zhou, Wenbo and Qin, Zheng},
  booktitle={Proceedings of the AAAI conference on artificial intelligence},
  volume={36},
  pages={951--959},
  year={2022}
}

@article{auxiliarybias,
  title={Auxiliary-loss-free load balancing strategy for mixture-of-experts},
  author={Wang, Lean and Gao, Huazuo and Zhao, Chenggang and Sun, Xu and Dai, Damai},
  journal={arXiv preprint arXiv:2408.15664},
  year={2024}
}

@article{multihead,
  title={Multi-head adapter routing for cross-task generalization},
  author={Page-Caccia, Lucas and Ponti, Edoardo Maria and Su, Zhan and Pereira, Matheus and Le Roux, Nicolas and Sordoni, Alessandro},
  journal={Advances in Neural Information Processing Systems},
  volume={36},
  pages={56916--56931},
  year={2023}
}

\clearpage
\appendix
\section{Appendix}

\subsection{A.More ablation studies}

In this section, we conduct cross-dataset ablation experiments on the DeepFakeBench benchmark to evaluate the effectiveness of our current hyperparameter settings.

1)the weighting coefficients of the multi-task loss.
In our framework, we introduce an authenticity classification branch and a forgery-type classification branch to encourage DFF-Adapter to capture fine-grained artifact cues in face forgery images. To assess the effectiveness of our loss weighting strategy, we perform ablation studies with different configurations of the multi-task loss coefficients. As shown in Table~\ref{tab:abla_lambda}, our selected setting achieves the highest AUC, demonstrating that the proposed balance between the two loss terms effectively enhances the model’s authenticity awareness.

2)Effect of Head Adapter Number.
To determine the optimal configuration of the Head Adapter number, we conduct a systematic ablation study by varying the channel partition count $h$ in the DFF-Adapter. As shown in Table~\ref{tab:abla_head}, a moderate number of Head Adapters achieves the best trade-off between parameter efficiency and feature diversity. Too few heads limit the independent modeling of forgery artefacts within subspaces, while an excessive number causes overfitting to local artifact cues. The selected configuration attains the highest AUC and most stable training performance across datasets, validating the rationality and effectiveness of this hyperparameter choice.

3)To further analyze the effect of parameter allocation within each head adapter, we conduct ablation experiments on the number of LoRA per head, with values of 1, 4, 6, and 12. Results show that moderate configurations achieve the best trade-off between model capacity and generalization.

\subsection{B.Grad-CAM Visualization of Attention}
To better understand the decision-making process of our approach, we present Grad-CAM visualizations on the CDF2, DFDC, and DFDCP datasets. As shown in Figure~\ref{fig:cam}, we further compare the visual explanations of our method with those of the DINOv2 baseline. While the baseline model occasionally fails to attend to forgery artifacts during classification, our proposed DFF-Adapter consistently focuses on the manipulated local regions of input faces, effectively highlighting tampered areas and exhibiting superior interpretability.

\begin{table}[t]
  \centering
  \setlength{\tabcolsep}{3pt}

  \begin{minipage}{\columnwidth}
    \centering
    \footnotesize
    \renewcommand{\arraystretch}{1.08}
    \begin{tabular}{@{}l l cc cc cc@{}}
      \toprule
    \multirow{2}{*}{$\lambda_0$} & \multirow{2}{*}{$\lambda_1$}   & \multicolumn{2}{c}{FF++} & \multicolumn{2}{c}{CDF-v2} & \multicolumn{2}{c}{DFDC} \\
      \cmidrule(lr){3-4}\cmidrule(lr){5-6}\cmidrule(lr){7-8}
      & & AUC & EER & AUC & EER & AUC & EER \\
      \midrule
      10 & 0  & 99.65 & 1.63 & 90.93 & 16.10 & 84.44 & 22.60 \\
      10 & 2  & 99.56 & 1.79 & \textbf{95.26} & \textbf{12.65} & \textbf{89.96} & \textbf{17.96} \\
      10 & 10 & \textbf{99.66} & \textbf{1.27} & 94.27 & 14.51 & 84.87 & 23.17 \\
      10 & 50 & 99.47 & 1.63 & 89.22 & 19.26 & 81.50 & 26.38 \\
      \bottomrule
    \end{tabular}
    \caption{\textbf{Ablation on the weighting coefficient $\lambda$ for the DFF-Adapter.} }
    \label{tab:abla_lambda}
  \end{minipage}

  \vspace{6pt}

  \begin{minipage}{\columnwidth}
    \centering
    \footnotesize
    \renewcommand{\arraystretch}{1.08}
    \begin{tabular}{@{}l cc cc cc@{}}
      \toprule
      \multirow{2}{*}{$h$} & \multicolumn{2}{c}{FF++} & \multicolumn{2}{c}{CDF-v2} & \multicolumn{2}{c}{DFDC} \\
      \cmidrule(lr){2-3}\cmidrule(lr){4-5}\cmidrule(lr){6-7}
      & AUC & EER & AUC & EER & AUC & EER \\
      \midrule
      1 & 99.22 & 1.18 & 90.95 & 16.68 & 82.47 & 24.40 \\
      4 & \textbf{99.56} & \textbf{1.79} & \textbf{95.26} & \textbf{12.65} & \textbf{89.96} & \textbf{17.96} \\
      8 & 99.40 & 1.81 & 94.58 & 12.21 & 84.38 & 22.43 \\
      \bottomrule
    \end{tabular}
    \caption{\textbf{Ablation on the number of head adapters ($h$) in the DFF-Adapter.}}
    \label{tab:abla_head}
  \end{minipage}

  \vspace{6pt}

  \begin{minipage}{\columnwidth}
    \centering
    \footnotesize
    \renewcommand{\arraystretch}{1.08}
    \begin{tabular}{@{}l cc cc cc@{}}
      \toprule
      \multirow{2}{*}{$n_L$} & \multicolumn{2}{c}{FF++} & \multicolumn{2}{c}{CDF-v2} & \multicolumn{2}{c}{DFDC} \\
      \cmidrule(lr){2-3}\cmidrule(lr){4-5}\cmidrule(lr){6-7}
      & AUC & EER & AUC & EER & AUC & EER \\
      \midrule
      1  & 99.57 & 1.99 & 93.73 & 15.38 & 84.31 & 23.32 \\
      4  & \textbf{99.69} & 2.08 & 91.85 & 17.24 & 83.81 & 23.68 \\
      6  & 99.56 & \textbf{1.79} & \textbf{95.26} & \textbf{12.65} & \textbf{89.96} & \textbf{17.96} \\
      12 & 99.52 & 1.83 & 90.24 & 18.68 & 82.60 & 24.92 \\
      \bottomrule
    \end{tabular}
   \caption{\textbf{Ablation on the number of LoRA experts per head ($n_L$).}}
    \label{tab:abla_loraperhead}
  \end{minipage}
\end{table}
\subsection{C.More Results in DF40 Dataset}
Beyond forgery detection, we further assess the capability of our method in identifying entirely synthesized face images generated by GANs and Diffusion models. To this end, we select eight representative generative methods from the DF40 dataset, including DiT-XL/2, RDDM, PixArt-$\alpha$, SiT-XL/2, StyleGAN3, StyleGAN-XL, and VQGAN, covering both GAN-based and diffusion-based Entire Face Synthesis approaches. As reported in Table~\ref{tab:entireface}, our method consistently achieves superior AUC performance compared to SBI, LVLM-DFD, and the DINOv2 extreme baseline. Notably, despite not being explicitly designed for this setting, our approach exhibits strong generalization ability and significantly outperforms existing methods on both GAN- and diffusion-based synthesis detection tasks.
\begin{figure*}[t]
\centering
\setlength{\tabcolsep}{2pt}
\renewcommand{\arraystretch}{1.0}
\begin{tabular}{>{\centering\arraybackslash}p{0.2cm} *{9}{c}}
  & \multicolumn{3}{c}{CDF2} & \multicolumn{3}{c}{DFDC} & \multicolumn{3}{c}{DFDCP} \\
\rotatebox{90}{Image}  & \includegraphics[width=0.08\textwidth]{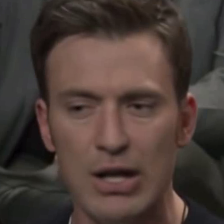} & \includegraphics[width=0.08\textwidth]{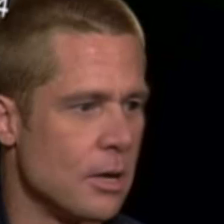} & \includegraphics[width=0.08\textwidth]{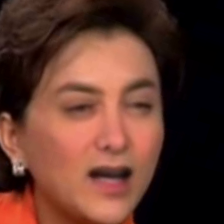} 
  & \includegraphics[width=0.08\textwidth]{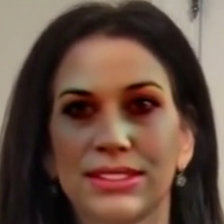} & \includegraphics[width=0.08\textwidth]{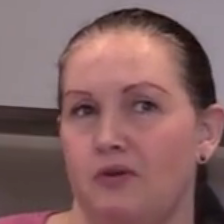} & \includegraphics[width=0.08\textwidth]{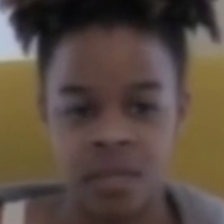}
  & \includegraphics[width=0.08\textwidth]{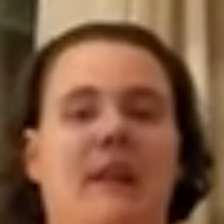} & \includegraphics[width=0.08\textwidth]{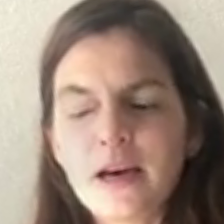} & \includegraphics[width=0.08\textwidth]{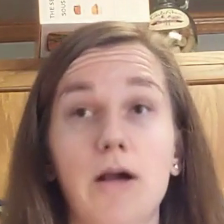} \\
\rotatebox{90}{DINOv2} & \includegraphics[width=0.08\textwidth]{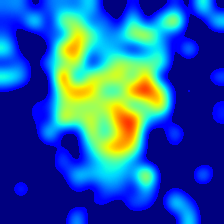} & \includegraphics[width=0.08\textwidth]{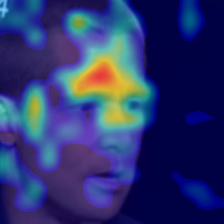} & \includegraphics[width=0.08\textwidth]{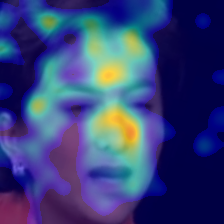} 
  & \includegraphics[width=0.08\textwidth]{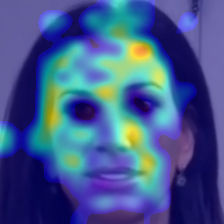} & \includegraphics[width=0.08\textwidth]{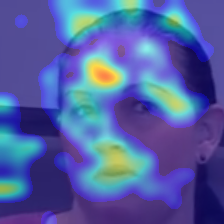} & \includegraphics[width=0.08\textwidth]{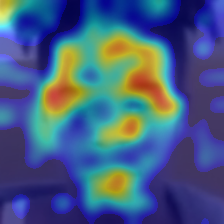}
  & \includegraphics[width=0.08\textwidth]{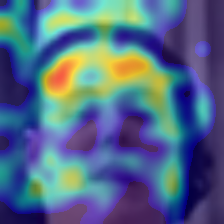} & \includegraphics[width=0.08\textwidth]{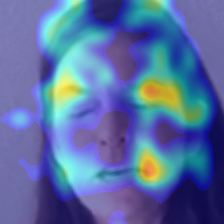} & \includegraphics[width=0.08\textwidth]{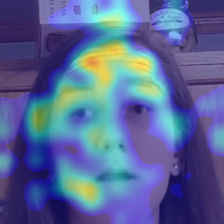} \\
\rotatebox{90}{Ours}   & \includegraphics[width=0.08\textwidth]{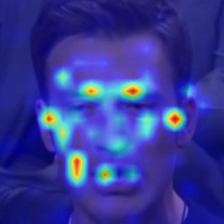} & \includegraphics[width=0.08\textwidth]{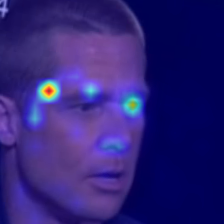} & \includegraphics[width=0.08\textwidth]{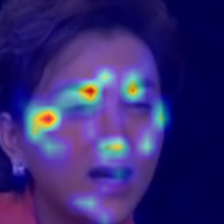} 
  & \includegraphics[width=0.08\textwidth]{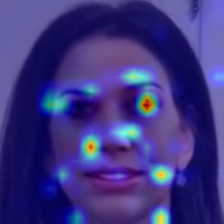} & \includegraphics[width=0.08\textwidth]{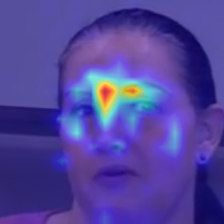} & \includegraphics[width=0.08\textwidth]{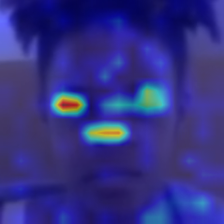}
  & \includegraphics[width=0.08\textwidth]{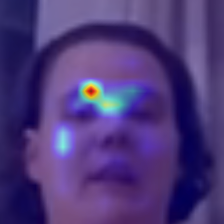} & \includegraphics[width=0.08\textwidth]{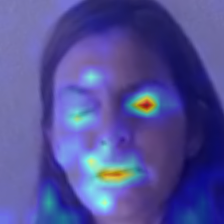} & \includegraphics[width=0.08\textwidth]{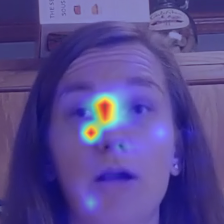} \\
\end{tabular}
\caption{Grad-CAM visualizations on the CDF2, DFDC, and DFDCP datasets. Our method produces more focused and interpretable attention maps compared to the DINOv2 baseline.}
\label{fig:cam}
\end{figure*}

\begin{table*}[t]
\centering
\small
\begin{tabular}{lccccccc}
\toprule
Method & DiT-XL/2 & RDDM & PixArt-$\alpha$ & SiT-XL/2 & StyleGAN3 & StyleGAN-XL & VQGAN \\
\midrule
SBI       & 79.04 & 53.66 & 98.78 & 84.27 & 97.91 & 23.26 & 91.50 \\
DINOv2    & 74.54 & 93.18 & 99.54&  75.25 & 85.93 & 87.70 &  89.02 \\
LVLM-DFD  & \textbf{94.18} & 86.61 & 100.00 & 94.12 & 98.87 & 100.00 & 99.99 \\
Ours      & 92.31 & \textbf{98.44} & 100.00 & \textbf{94.65} & \textbf{99.92} & 100.00 & \textbf{100.00} \\
\bottomrule
\end{tabular}
\caption{\textbf{Performance on GAN and diffusion-based Entire Face Synthesis methods from the DF40 dataset.}}
\label{tab:entireface}
\end{table*}
\subsection{D.Limitation}
While our method leverages forgery-type classification as an auxiliary task to enhance authenticity discrimination, it inherently assumes that different manipulation methods produce distinct and consistent artifact patterns. However, in real-world scenarios, such an assumption may not always hold. The artifacts introduced by different forgery techniques can be partially overlapping, subtle, or evolve with newer generation methods, leading to ambiguity in manipulation-type classification. This overlap may introduce noise into the auxiliary task and, in turn, affect the main authenticity detection stream by propagating less reliable gradients. Moreover, some advanced forgery techniques may intentionally mimic authentic distributions or blend characteristics of multiple manipulation types, making it increasingly difficult to disentangle class-specific cues. These factors could limit the effectiveness of the forgery-type classification branch and potentially reduce the overall generalization of the detection model.

\end{document}